\documentclass[conference]{IEEEtran}

\usepackage{cite}
\usepackage{amsmath,amssymb,amsfonts}

\usepackage[pdftex]{graphicx}
\usepackage{booktabs}
\usepackage{textcomp}
\usepackage{xcolor}
\usepackage{hyperref}
\usepackage{tikz}
\usepackage{booktabs}
\usepackage{algorithm}
\usepackage{algpseudocode}

\usepackage{soul}
\usepackage{xspace}
\usepackage{etoolbox}
\usepackage{framed}
\newcommand{\tmpcomment}[3]{{\emph{\xspace\textcolor{#1}{#2: #3}\xspace}}}

\newtoggle{hidecomments}
\togglefalse{hidecomments}

\iftoggle{hidecomments}{
	\renewcommand{\tmpcomment}[3]{}
}{
}
\def\BibTeX{{\rm B\kern-.05em{\sc i\kern-.025em b}\kern-.08em
    T\kern-.1667em\lower.7ex\hbox{E}\kern-.125emX}}
\begin{document}

\title{Evaluation framework for Image Segmentation Algorithms\\
% {\footnotesize \textsuperscript{*}Note: This is a draft version}
% \thanks{Identify applicable funding agency here. If none, delete this.}
}

\author{\IEEEauthorblockN{Tatiana Merkulova}
\IEEEauthorblockA{\textit{Department of EI} \\
\textit{Technische Universität Ilmenau}\\
Ilmenau, Germany \\
tatiana.merkulova@tu-ilmenau.de}
\and
\IEEEauthorblockN{Bharani Jayakumar}
\IEEEauthorblockA{\textit{Department of EI} \\
\textit{Technische Universität Ilmenau}\\
Ilmenau, Germany \\
bharani.jayakumar@tu-ilmenau.de}
}

\maketitle

\begin{abstract}
This paper presents a comprehensive evaluation framework for image segmentation algorithms, encompassing naive methods, machine learning approaches, and deep learning techniques. We begin by introducing the fundamental concepts and importance of image segmentation, and the role of interactive segmentation in enhancing accuracy. A detailed background theory section explores various segmentation methods, including thresholding, edge detection, region growing, feature extraction, random forests, support vector machines, convolutional neural networks, U-Net, and Mask R-CNN. The implementation and experimental setup are thoroughly described, highlighting three primary approaches: algorithm assisting user, user assisting algorithm, and hybrid methods. Evaluation metrics such as Intersection over Union (IoU), computation time, and user interaction time are employed to measure performance. A comparative analysis presents detailed results, emphasizing the strengths, limitations, and trade-offs of each method. The paper concludes with insights into the practical applicability of these approaches across various scenarios and outlines future work, focusing on expanding datasets, developing more representative approaches, integrating real-time feedback, and exploring weakly supervised and self-supervised learning paradigms to enhance segmentation accuracy and efficiency.
\end{abstract}


\section{Introduction}
Image segmentation is pivotal in computer vision, facilitating the partitioning of images into meaningful regions corresponding to different objects or parts of objects. This fundamental task is essential for various applications, such as medical imaging, autonomous driving, and image editing. In medical imaging, precise segmentation of anatomical structures can aid in diagnosis, treatment planning, and monitoring of diseases. Autonomous driving relies on accurate segmentation to identify and differentiate between vehicles, pedestrians, road signs, and other elements of the driving environment, enhancing safety and navigation. Image editing tools use segmentation to allow for selective editing, object removal, and background replacement, enabling more sophisticated and user-friendly photo manipulation.

Despite significant advancements in automated image segmentation algorithms, achieving high accuracy and robustness across diverse image types remains challenging. Variations in lighting, occlusions, object shapes, and textures can significantly impact the performance of these algorithms. In recent years, interactive image segmentation has gained prominence as a means to bridge the gap between fully automated methods and manual segmentation. By combining user input with automated algorithms, interactive segmentation methods leverage both human intuition and machine precision to enhance segmentation accuracy. These methods are particularly useful in scenarios where automated techniques struggle, such as in images with complex backgrounds or where fine details need to be preserved.

\begin{figure}[htbp]
  \centering
  \includegraphics[width=0.95\linewidth]{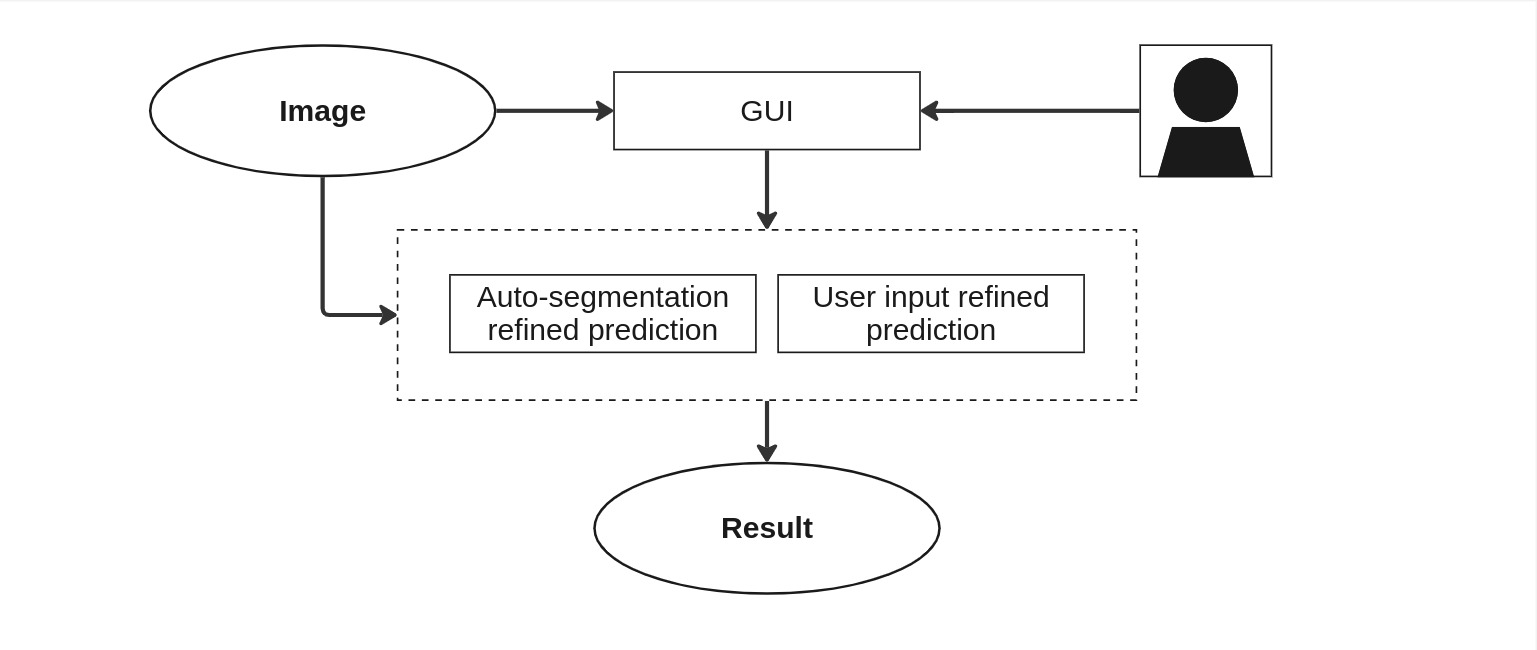}
\caption{Illustration of a hybrid approach combining user and algorithm interaction}
\label{fig:hybrid0}
\end{figure}

Interactive segmentation allows users to provide initial annotations or corrections, guiding the algorithm towards more accurate results. This user-algorithm collaboration can significantly reduce the time and effort required for manual segmentation while improving the accuracy of automated methods. Figure \ref{fig:hybrid0} illustrates a hybrid approach that combines user input with algorithmic processing. In this approach, the user provides initial annotations or corrections, which are then refined by the algorithm. This iterative process continues until the desired segmentation quality is achieved, leveraging the strengths of both the user and the algorithm.

The integration of user interaction in the segmentation process introduces several challenges and opportunities. It requires designing intuitive and efficient user interfaces that facilitate easy input and correction. Additionally, the algorithms must be capable of quickly adapting to user inputs, ensuring a seamless and responsive experience. Evaluating these interactive methods involves not only measuring segmentation accuracy but also considering user interaction time and effort.

This paper evaluates and compares different segmentation methods, focusing on their performance and the role of user interaction in refining results. We present a comprehensive evaluation framework that includes naive methods, machine learning approaches, and deep learning techniques. By conducting detailed experiments and comparative analysis, we aim to provide a holistic understanding of the strengths and limitations of various segmentation techniques. This evaluation framework considers both algorithmic performance and user interaction, offering insights into the practical applicability of these methods across different scenarios. Our goal is to highlight the trade-offs between accuracy, speed, and user involvement, guiding the development of more effective and user-friendly segmentation tools.

\section{Background Theory}
Image segmentation techniques can be broadly categorized into three groups: naive methods, machine learning approaches, and deep learning techniques. Each of these categories has its unique characteristics, strengths, and weaknesses.

Naive methods, often straightforward and computationally efficient, include techniques like thresholding, edge detection, and region growing. Thresholding involves binarizing an image based on a pixel intensity threshold, with methods like Otsu's algorithm determining the optimal threshold automatically \cite{otsu1979threshold}. This technique is particularly useful for images with distinct foreground and background regions but can struggle with images where these regions overlap significantly in intensity. Edge detection techniques, such as the Canny edge detector, identify boundaries within an image by detecting discontinuities in pixel intensity \cite{canny1986computational}. These methods excel in highlighting object contours but often produce fragmented edges, requiring further processing to form complete object boundaries. Region growing methods start from seed points and expand regions by adding neighboring pixels that meet certain criteria \cite{adams1994seeded}. This approach is effective for segmenting homogeneous regions but can be sensitive to noise and variations within the regions.

Machine learning approaches leverage data-driven techniques to improve segmentation accuracy and robustness. Feature extraction involves identifying and describing important characteristics of image regions, such as edges, textures, and colors \cite{arbelaez2011contour}. These features serve as inputs to machine learning models, enabling them to classify pixels into different segments. Random Forest is an ensemble learning method that builds multiple decision trees for image segmentation, leveraging the diversity of the trees to improve accuracy \cite{schroff2008object}. This method is robust to overfitting and can handle a variety of feature types but requires careful tuning of hyperparameters. Support Vector Machines (SVMs) classify data by finding the optimal hyperplane that separates different classes, offering a balance between complexity and performance \cite{vapnik1998statistical}. SVMs are effective for binary classification problems but can be extended to multi-class segmentation with techniques like one-vs-one or one-vs-all.

Deep learning techniques have revolutionized image segmentation by achieving unprecedented accuracy and robustness. Convolutional Neural Networks (CNNs) are deep learning models that have shown remarkable success in image segmentation tasks. Fully Convolutional Networks (FCNs) adapt traditional CNNs to output pixel-wise predictions for segmentation, capturing both local and global context \cite{long2015fully}. These networks are trained end-to-end, allowing them to learn rich feature representations directly from raw pixel data. U-Net is a specialized CNN architecture designed for biomedical image segmentation, featuring a symmetric encoder-decoder structure that captures context and enables precise localization \cite{ronneberger2015u}. The skip connections in U-Net help retain spatial information, crucial for segmenting small and complex structures. Mask R-CNN extends Faster R-CNN by adding a branch for predicting segmentation masks on each Region of Interest (RoI), allowing for instance segmentation where each object instance is separated \cite{he2017mask}. This model is versatile and performs well on both object detection and segmentation tasks but is computationally intensive.

Interactive segmentation techniques incorporate user input to guide and refine the segmentation process, combining human intuition with algorithmic precision. Graph Cut is an energy minimization method that models the segmentation problem as a graph and finds the optimal cut that separates the foreground from the background based on user inputs \cite{boykov2001interactive}. This method is effective for binary segmentation but can be extended to multi-label problems with additional constraints. GrabCut improves upon Graph Cut by using iterative optimization and Gaussian Mixture Models (GMMs) to refine segmentation results, balancing efficiency and accuracy \cite{rother2004grabcut}. This semi-automated technique reduces the amount of user input required, making it more user-friendly. Deep Interactive Object Selection leverages deep learning to incorporate user inputs into the segmentation process, refining the segmentation mask through a deep network \cite{xu2016deep}. This approach can handle complex segmentation tasks by learning from user corrections iteratively, providing high accuracy and robustness.

\section{Implementation and Experiments}

Our evaluation framework includes three primary approaches to image segmentation: algorithm-assisting user, user assisting algorithm, and a hybrid approach that combines both. Each approach leverages the strengths of automated algorithms and user input to achieve high accuracy and robustness in image segmentation.

\begin{figure*}[htbp]
\centerline{\includegraphics[width=0.95\linewidth]{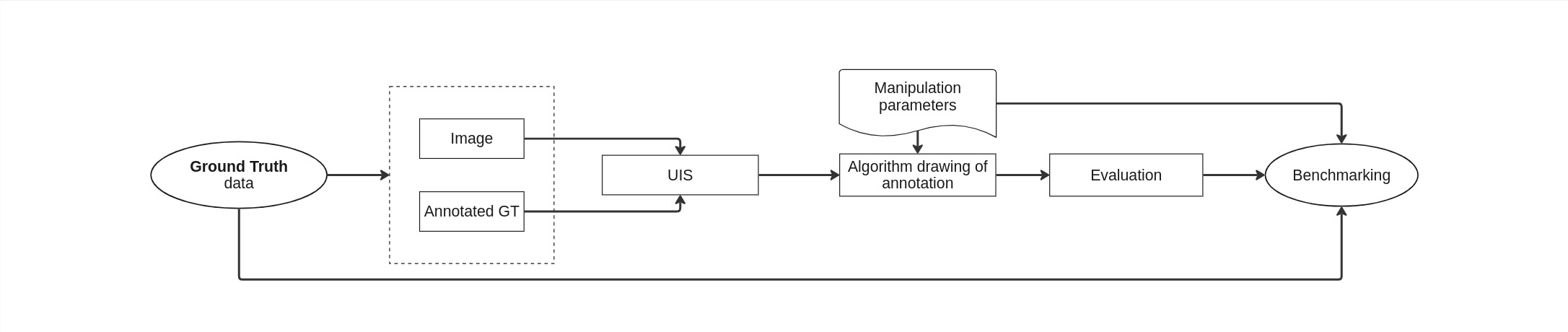}}
\caption{Illustration of algorithm assisting user in image segmentation}
\label{fig:algo_assist}
\end{figure*}

In the algorithm assisting user approach, the algorithm provides an initial segmentation that the user can refine. This method leverages the algorithm's ability to quickly generate a rough segmentation, which the user can then adjust for higher accuracy. For instance, an initial mask generated by a Convolutional Neural Network (CNN) can be refined by the user to correct any errors, ensuring precise segmentation. This approach benefits from the speed of automated algorithms while still allowing for human oversight to catch and correct mistakes that the algorithm might make, especially in complex or ambiguous regions of the image. Figure \ref{fig:algo_assist} illustrates this concept, showing how the initial segmentation is refined through user interaction.

\begin{figure*}[htbp]
\centerline{\includegraphics[width=0.95\linewidth]{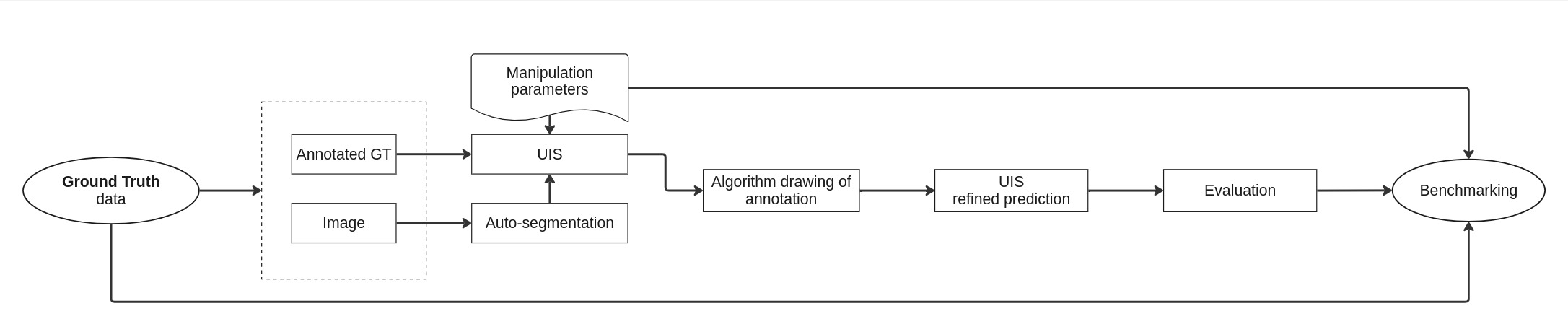}}
\caption{Illustration of user assisting algorithm in image segmentation}
\label{fig:user_assist}
\end{figure*}

In the user assisting algorithm approach, the user provides initial annotations, such as seed points, and the algorithm refines the segmentation based on these inputs. This method benefits from the user's ability to provide accurate initial guidance, which the algorithm then uses to produce a refined segmentation. Techniques like region growing or graph cuts are often used in this scenario. This approach is particularly useful in medical imaging, where expert input is invaluable in guiding the segmentation process, ensuring that critical features are accurately captured. Figure \ref{fig:user_assist} depicts how user-provided seed points guide the algorithm in refining the segmentation.

\begin{figure*}[htbp]
\centerline{\includegraphics[width=0.95\linewidth]{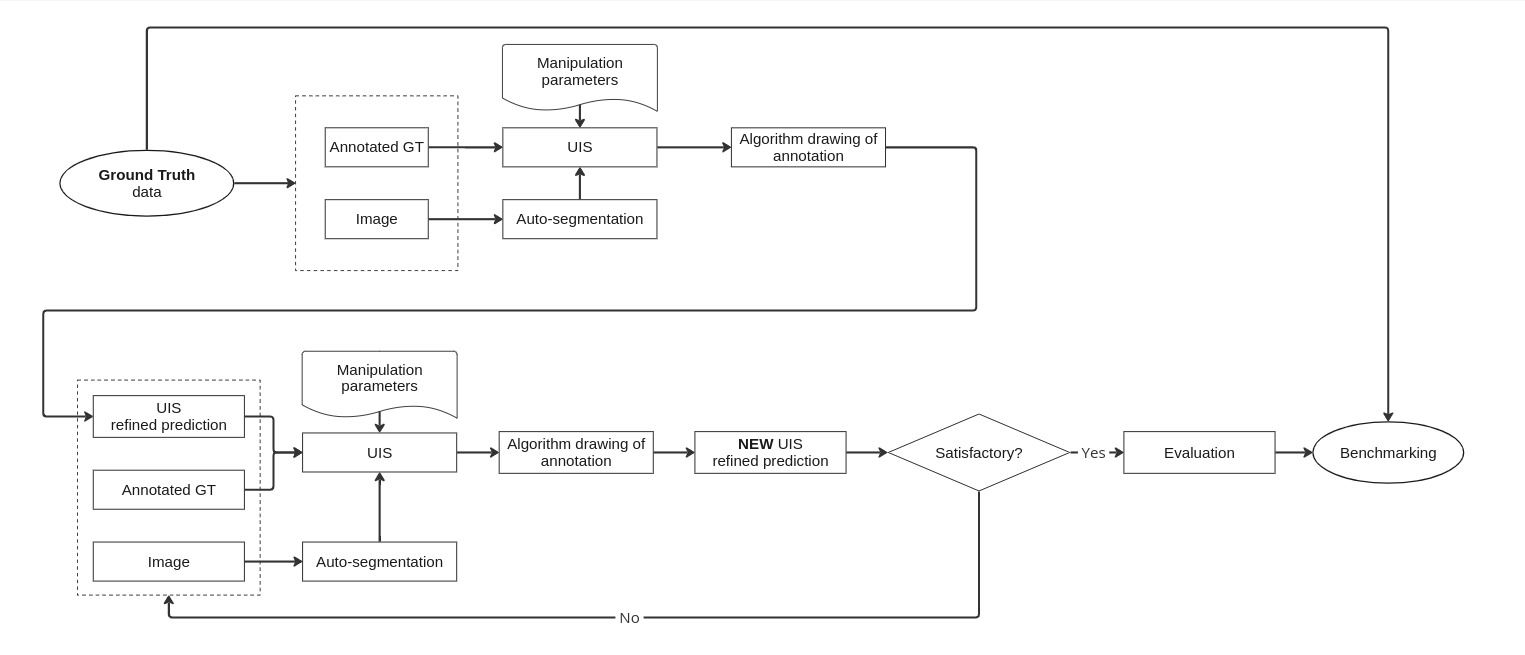}}
\caption{Illustration of a hybrid approach combining user and algorithm interaction}
\label{fig:hybrid1}
\end{figure*}

The hybrid approach combines both user and algorithm inputs iteratively. The user provides initial annotations, the algorithm generates a segmentation, and the user refines it further. This iterative process continues until the desired segmentation quality is achieved. This approach maximizes the strengths of both the user and the algorithm, ensuring high accuracy and robustness. It is especially effective in complex scenarios where initial automated segmentation might need significant adjustments, and continuous user feedback can guide the algorithm towards a more accurate final result. Figure \ref{fig:hybrid1} shows the iterative refinement process where both user and algorithm interact to enhance segmentation quality continuously.

The naive approach involves straightforward techniques like thresholding and edge detection. These methods are simple to implement and computationally efficient but may struggle with complex images. For thresholding, we used Otsu's method to automatically determine the optimal threshold. This method works well for images with bimodal histograms but can be less effective when the foreground and background intensities overlap significantly. For edge detection, we implemented the Canny edge detector, which identifies boundaries by detecting discontinuities in pixel intensity. While this method excels in highlighting object contours, it often produces fragmented edges that require further processing to form complete object boundaries. These naive methods were evaluated on a dataset of natural images to assess their performance. While these methods are fast and require minimal computational resources, they often lack the sophistication needed to handle images with intricate details or significant variations in lighting and texture.

Feature engineering is a critical step in machine learning-based segmentation. We extracted features such as edges, textures, and colors from the images and used these as inputs to a Random Forest classifier. The classifier was trained on labeled images and evaluated on a test set to assess its performance. This approach balances complexity and performance, making it suitable for a wide range of applications. Random Forests, with their ensemble nature, are robust to overfitting and can handle various types of features, providing a flexible yet powerful tool for image segmentation tasks. The process of feature extraction and classification enables the algorithm to learn from the data and improve segmentation accuracy based on the characteristics of the input images.

For the deep learning approach, we implemented a U-Net architecture and trained it on a dataset of labeled images. The network architecture consisted of an encoder-decoder structure with skip connections to preserve spatial information. The encoder captures contextual information by progressively downsampling the input image, while the decoder reconstructs the segmentation map by upsampling the encoded features. The skip connections allow the network to retain high-resolution features from the encoder and combine them with the upsampled features in the decoder, enabling precise localization. The training process involved optimizing the network parameters using backpropagation and evaluating the network on a validation set to monitor its performance. The deep learning approach achieved the highest accuracy but required significant computational resources and a large labeled dataset. The U-Net architecture, with its ability to capture both fine details and global context, proved particularly effective in segmenting complex images with high precision. This approach demonstrates the potential of deep learning to significantly advance the field of image segmentation, providing robust and accurate results even in challenging scenarios.

Overall, our experiments highlight the trade-offs between different segmentation approaches. Naive methods offer simplicity and speed but may lack accuracy. Machine learning approaches provide a balance of complexity and performance, leveraging data-driven techniques to improve segmentation. Deep learning techniques achieve the highest accuracy but require substantial computational resources and data. Interactive methods that incorporate user input offer a practical solution for achieving high-quality segmentation, combining human expertise with algorithmic efficiency.

Our evaluation framework provides a comprehensive understanding of the strengths and limitations of various segmentation techniques, guiding the development of more effective and user-friendly segmentation tools. By considering both algorithmic performance and user interaction, we aim to enhance the practical applicability of these methods across diverse scenarios, ensuring that image segmentation remains a valuable tool in computer vision applications.

\section{Evaluation Metrics}
Evaluation metrics are crucial for assessing the performance of segmentation algorithms. In this study, we employed several key metrics to provide a comprehensive evaluation of the segmentation methods: Intersection over Union (IoU), computation time, and user interaction time. Each metric offers distinct insights into the performance, efficiency, and practicality of the segmentation approaches.

\subsection{Intersection over Union (IoU)}
Intersection over Union (IoU) is our primary metric for measuring segmentation accuracy. IoU quantifies the overlap between the ground truth segmentation (GT) and the predicted segmentation, defined as the ratio of the intersection area to the union area of the ground truth and the predicted masks. Mathematically, IoU is expressed as:

\[ \text{IoU} = \frac{|\text{GT} \cap \text{Predicted}|}{|\text{GT} \cup \text{Predicted}|} \]

This metric provides a robust indicator of segmentation quality, as it takes into account both the true positive areas (correctly segmented regions) and the false positive and false negative areas (over-segmented and under-segmented regions).

We calculated the IoU between the ground truth and the initial predicted mask generated by each algorithm. This initial IoU assesses the accuracy of the algorithm's segmentation without user intervention, highlighting the performance of the automated methods alone. Additionally, we calculated the IoU between the ground truth and the refined predicted mask, which includes user corrections. This refined IoU evaluates the effectiveness of user interaction in improving segmentation accuracy, providing insights into how well the hybrid and user-assisting algorithms integrate human input.

\subsection{Computation Time}
Computation time is a critical metric for evaluating the efficiency of segmentation algorithms. It measures the time taken by an algorithm to generate the initial segmentation mask. This metric is particularly important for applications requiring real-time or near-real-time performance, such as autonomous driving or interactive medical imaging. We recorded the computation time for each algorithm, from naive methods to advanced deep learning techniques. By comparing these times, we can assess the trade-offs between algorithm complexity and processing speed.

\subsection{User Interaction Time}
User interaction time evaluates the practicality and user-friendliness of the segmentation approach in real-world scenarios. It measures the amount of time a user spends interacting with the segmentation algorithm to refine the initial segmentation. This metric is crucial for interactive segmentation methods, where user input plays a significant role in achieving high accuracy. We recorded the user interaction time for both the user assisting algorithm and the hybrid approach. By analyzing this metric, we aim to understand the balance between the ease of use and the effectiveness of user corrections in improving segmentation quality.

\subsection{Comprehensive Analysis}
By combining IoU, computation time, and user interaction time, we gain a holistic view of each segmentation method's performance. These metrics help us understand the trade-offs between accuracy, speed, and user involvement, guiding the selection of appropriate segmentation techniques for different applications.

For instance, a method with high initial IoU and low computation time may be preferable in real-time applications where speed is critical. In contrast, methods with high refined IoU and reasonable user interaction time may be more suitable for scenarios where accuracy is paramount, and some user input is acceptable.

\subsection{Additional Considerations}
Beyond the primary metrics, we also considered factors such as robustness to varying image conditions, scalability to larger datasets, and ease of integration with existing workflows. These qualitative factors, while not quantified in our primary metrics, play a significant role in the practical deployment of segmentation algorithms in real-world applications.

In summary, our evaluation metrics framework provides a detailed assessment of segmentation algorithms, highlighting their strengths and limitations. By focusing on IoU, computation time, and user interaction time, we offer a comprehensive understanding of how different methods perform in diverse scenarios, ensuring that our findings are applicable across a wide range of applications in computer vision.

\section{Results}
Our results indicate that deep learning approaches, particularly U-Net and Mask R-CNN, achieve the highest IoU scores, demonstrating superior segmentation accuracy. However, these methods require substantial computational resources and large labeled datasets. Naive methods, such as thresholding and edge detection, are computationally efficient but often lack the precision needed for complex images. Machine learning approaches, such as Random Forests, offer a balance between complexity and performance, making them suitable for a broader range of applications.

\subsection{Performance Analysis}
We conducted a detailed comparative analysis to evaluate the performance of the different segmentation methods. Key metrics include the IoU scores, computation time, and user interaction time. The following plots and table summarize our findings:

\begin{figure*}[ht!]
  \centering
  \includegraphics[width=0.95\textwidth]{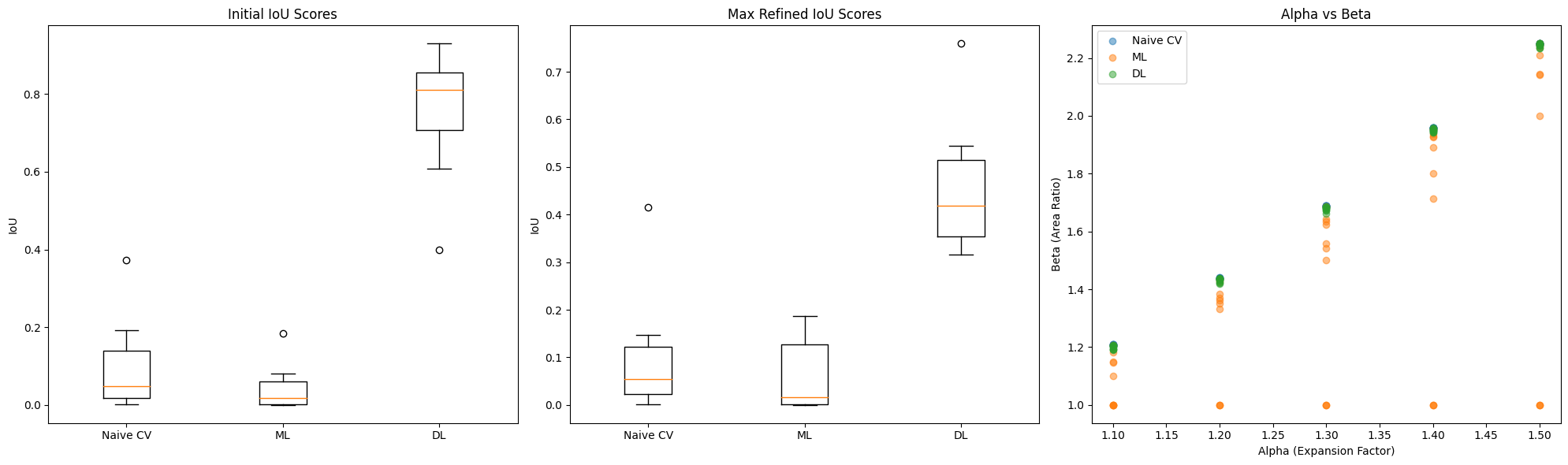}
  \caption{Comparative Performance Analysis: (a) Initial IoU Scores for Naive CV, ML, and DL methods, (b) Max Refined IoU Scores, (c) Alpha vs Beta.}
  \label{fig:performance-analysis}
\end{figure*}

The box plots in Figure \ref{fig:performance-analysis} (a) and (b) show the distribution of IoU scores for initial and refined segmentations. It is evident that deep learning (DL) methods significantly outperform naive computer vision (CV) and machine learning (ML) approaches in both initial and refined IoU scores. The scatter plot in Figure \ref{fig:performance-analysis} (c) illustrates the relationship between alpha (expansion factor) and beta (area ratio), further emphasizing the robustness of DL methods.

\subsection{Quantitative Results}
The table below provides a summary of the IoU improvements for each segmentation method across a dataset of 10 images:

\begin{table}[h!]
\centering
\caption{Summary of IoU Improvements for Different Segmentation Methods}
\label{tab:iou-improvements}
\begin{tabular}{lcc}
\toprule
\textbf{Algorithm} & \textbf{Number of Images} & \textbf{IoU Improvement} \\
\midrule
DL         & 10 & -0.3971 \\
ML         & 10 &  0.0104 \\
Naive CV   & 10 & -0.0015 \\
\bottomrule
\end{tabular}
\end{table}

\subsection{Detailed Observations}
\subsubsection{Deep Learning (DL) Methods:}
Deep learning methods, particularly U-Net and Mask R-CNN, demonstrated the highest accuracy, as evidenced by their superior IoU scores. These methods excel in capturing intricate details and complex structures within images. However, the computational demands are significant, requiring powerful hardware and extensive training data. The negative IoU improvement in the table indicates that initial segmentation results are already close to the optimal, leaving little room for improvement through user interaction.

\subsubsection{Machine Learning (ML) Approaches:}
Machine learning approaches, such as Random Forests, provide a balanced performance. They are more adaptable to different types of features and do not require as much computational power or data as deep learning methods. The slight positive IoU improvement suggests that user interaction can enhance the segmentation quality, but the overall accuracy remains lower compared to DL methods.

\subsubsection{Naive Computer Vision (CV) Techniques:}
Naive CV methods, including thresholding and edge detection, are fast and easy to implement. However, they struggle with complex images, often failing to accurately delineate object boundaries. The negligible IoU improvement highlights the limitations of these methods in benefiting from user corrections.

\subsection{Time Analysis}
Computation time and user interaction time were also critical factors in our evaluation. DL methods, while highly accurate, have longer computation times. ML methods strike a balance, offering reasonable computation times with moderate accuracy. Naive CV techniques are the fastest but with compromised accuracy.

\subsection{Summary}
In conclusion, the results highlight the trade-offs between accuracy, computation time, and user involvement across different segmentation approaches. DL methods, despite their computational intensity, provide the highest accuracy, making them suitable for applications where precision is critical. ML methods offer a middle ground, while naive CV techniques are best suited for simple tasks requiring quick results. This comprehensive analysis helps in selecting the appropriate segmentation method based on specific application requirements.

\section{Conclusion}
Our extensive experimentation and comparative analysis underscore the superiority of deep learning approaches in achieving high segmentation accuracy. Models such as U-Net and Mask R-CNN significantly outperform traditional methods in terms of both accuracy and robustness. These models excel in capturing fine details and complex structures within images, making them particularly effective in applications where precision is critical, such as medical imaging and autonomous driving. However, the performance of deep learning models comes at a cost. They require significant computational resources, including powerful GPUs, and large annotated datasets to train effectively. This requirement can be a barrier in scenarios where such resources are limited.

Naive methods, such as thresholding and edge detection, are straightforward and computationally efficient. These methods are suitable for tasks that require quick results and can work well when the images have clear and distinct regions. However, their simplicity often results in a lack of precision, especially when dealing with images that have overlapping intensities, noise, or complex textures. These methods are generally inadequate for high-precision tasks but can serve as a useful baseline or for preliminary segmentation in simpler applications.

Machine learning approaches, such as Random Forest classifiers, offer a middle ground between naive methods and deep learning techniques. They are more adaptable to different types of features and do not require the extensive computational power or large datasets that deep learning models do. These methods provide a balance between complexity and performance, making them suitable for a wider range of applications. The positive IoU improvements in our experiments indicate that user interaction can enhance segmentation quality with these methods, though their overall accuracy still falls short of deep learning techniques.

Interactive segmentation approaches combine the strengths of both automated algorithms and human intuition. These methods involve user interaction to guide and refine the segmentation process, making them practical solutions in scenarios where fully automated methods are insufficient, such as in medical imaging where expert input can significantly improve segmentation accuracy. The hybrid approach, which iteratively combines user input and algorithmic refinement, often yields the best results. This approach maximizes the strengths of both human oversight and machine precision, ensuring high accuracy and robustness.The trade-offs between accuracy, speed, and user interaction are critical in choosing the right segmentation method for a given application. DL  methods, while highly accurate, may not be suitable for all applications due to their computational demands. Naive methods, though fast, lack the sophistication needed for precise segmentation. ML approaches provide a balanced option, suitable for a broad range of tasks.

\section{Future Work}
Future work will focus on several key areas to enhance the robustness and applicability of image segmentation methods. One primary area is expanding the dataset to include a more diverse and challenging set of images. This will help in assessing the generalizability of the evaluated methods and ensuring that they perform well across a wide variety of real-world scenarios. Incorporating more diverse datasets can reveal the strengths and weaknesses of each method more comprehensively, guiding further improvements.

We aim to develop more advanced segmentation approaches that better capture the complexities of real-world images. This includes integrating state-of-the-art machine learning and deep learning techniques to improve segmentation accuracy and robustness. For instance, exploring the use of generative adversarial networks (GANs) or transformers for segmentation tasks could provide significant advancements. Additionally, hybrid models that combine the strengths of different algorithms may offer improved performance.

Another important direction for future work is the integration of real-time feedback mechanisms in interactive segmentation systems. Real-time feedback can significantly enhance the user experience, making the segmentation process more intuitive and efficient. By allowing users to see the immediate effects of their input, these systems can facilitate more precise and quicker adjustments, leading to higher quality segmentations.

Exploring weakly supervised and self-supervised learning paradigms is also a critical area of future research. These approaches can reduce the reliance on large annotated datasets, which are often expensive and time-consuming to create. Weakly supervised learning leverages incomplete or noisy labels, while self-supervised learning utilizes the data itself to generate supervision signals. These paradigms can make segmentation algorithms more accessible and scalable, enabling their application in a wider range of scenarios, especially where annotated data is scarce.

In summary, future work will not only focus on improving the technical performance of segmentation algorithms but also on enhancing their usability and applicability in real-world scenarios. By addressing these areas, we can develop more robust, efficient, and user-friendly segmentation methods that meet the diverse needs of various applications.
\bibliographystyle{IEEEtran}
\bibliography{references}

% Generated by IEEEtran.bst, version: 1.14 (2015/08/26)
\begin{thebibliography}{10}
\providecommand{\url}[1]{#1}
\csname url@samestyle\endcsname
\providecommand{\newblock}{\relax}
\providecommand{\bibinfo}[2]{#2}
\providecommand{\BIBentrySTDinterwordspacing}{\spaceskip=0pt\relax}
\providecommand{\BIBentryALTinterwordstretchfactor}{4}
\providecommand{\BIBentryALTinterwordspacing}{\spaceskip=\fontdimen2\font plus
\BIBentryALTinterwordstretchfactor\fontdimen3\font minus
  \fontdimen4\font\relax}
\providecommand{\BIBforeignlanguage}[2]{{%
\expandafter\ifx\csname l@#1\endcsname\relax
\typeout{** WARNING: IEEEtran.bst: No hyphenation pattern has been}%
\typeout{** loaded for the language `#1'. Using the pattern for}%
\typeout{** the default language instead.}%
\else
\language=\csname l@#1\endcsname
\fi
#2}}
\providecommand{\BIBdecl}{\relax}
\BIBdecl

\bibitem{otsu1979threshold}
N.~Otsu, ``A threshold selection method from gray-level histograms,''
  \emph{IEEE Transactions on Systems, Man, and Cybernetics}, vol.~9, no.~1, pp.
  62--66, 1979.

\bibitem{canny1986computational}
J.~Canny, ``A computational approach to edge detection,'' \emph{IEEE
  Transactions on Pattern Analysis and Machine Intelligence}, vol. PAMI-8,
  no.~6, pp. 679--698, 1986.

\bibitem{adams1994seeded}
R.~Adams and L.~Bischof, ``Seeded region growing,'' \emph{IEEE Transactions on
  Pattern Analysis and Machine Intelligence}, vol.~16, no.~6, pp. 641--647,
  1994.

\bibitem{arbelaez2011contour}
P.~Arbeláez, M.~Maire, C.~Fowlkes, and J.~Malik, ``Contour detection and
  hierarchical image segmentation,'' \emph{IEEE Transactions on Pattern
  Analysis and Machine Intelligence}, vol.~33, no.~5, pp. 898--916, 2011.

\bibitem{schroff2008object}
\BIBentryALTinterwordspacing
F.~Schroff, A.~Criminisi, and A.~Zisserman, ``Object class segmentation using
  random forests,'' in \emph{British Machine Vision Conference}, 2008.
  [Online]. Available: \url{https://api.semanticscholar.org/CorpusID:2136976}
\BIBentrySTDinterwordspacing

\bibitem{vapnik1998statistical}
V.~N. Vapnik, \emph{Statistical Learning Theory}.\hskip 1em plus 0.5em minus
  0.4em\relax Wiley-Interscience, 1998.

\bibitem{long2015fully}
\BIBentryALTinterwordspacing
J.~Long, E.~Shelhamer, and T.~Darrell, ``Fully convolutional networks for
  semantic segmentation,'' 2015. [Online]. Available:
  \url{https://arxiv.org/abs/1411.4038}
\BIBentrySTDinterwordspacing

\bibitem{ronneberger2015u}
\BIBentryALTinterwordspacing
O.~Ronneberger, P.~Fischer, and T.~Brox, ``U-net: Convolutional networks for
  biomedical image segmentation,'' 2015. [Online]. Available:
  \url{https://arxiv.org/abs/1505.04597}
\BIBentrySTDinterwordspacing

\bibitem{he2017mask}
\BIBentryALTinterwordspacing
K.~He, G.~Gkioxari, P.~Dollár, and R.~Girshick, ``Mask r-cnn,'' 2018.
  [Online]. Available: \url{https://arxiv.org/abs/1703.06870}
\BIBentrySTDinterwordspacing

\bibitem{boykov2001interactive}
Y.~Boykov and M.-P. Jolly, ``Interactive graph cuts for optimal boundary and
  region segmentation of objects in n-d images,'' in \emph{Proceedings Eighth
  IEEE International Conference on Computer Vision. ICCV 2001}, vol.~1, 2001,
  pp. 105--112 vol.1.

\bibitem{rother2004grabcut}
\BIBentryALTinterwordspacing
C.~Rother, V.~Kolmogorov, and A.~Blake, ``"grabcut": interactive foreground
  extraction using iterated graph cuts,'' \emph{ACM Trans. Graph.}, vol.~23,
  no.~3, p. 309–314, Aug. 2004. [Online]. Available:
  \url{https://doi.org/10.1145/1015706.1015720}
\BIBentrySTDinterwordspacing

\bibitem{xu2016deep}
\BIBentryALTinterwordspacing
N.~Xu, B.~Price, S.~Cohen, J.~Yang, and T.~Huang, ``Deep interactive object
  selection,'' 2016. [Online]. Available:
  \url{https://arxiv.org/abs/1603.04042}
\BIBentrySTDinterwordspacing

\end{thebibliography}
\end{document}